%% file: main.tex
\documentclass[letterpaper,journal]{IEEEtran}
\input{preamble}

\usepackage[colorlinks,
linkcolor=blue,
anchorcolor=blue,
citecolor=blue]{hyperref}
\title{Efficient Multi-View Generation with Iterative Global-Local Proxy Attention}
\author{Wei~Tan,
        Kunming~Luo,
        Hongyu~Yan,
        Yu~Zhang,
        Zihao~Zhang,
        Xiatian~Zhu,
        and~Lei~Xie%
\thanks{Wei Tan and Lei Xie are with the School of Cyber Science
and Engineering, Southeast University, Nanjing 210096, China.}%
\thanks{Kunming Luo and Hongyu Yan are with the Department of Electronic and Computer Engineering, the Hong Kong University of Science and Technology, Hong Kong, China.}%
\thanks{Yu Zhang is with the College of Electronic and Information Engineering, Southwest University, Chongqing, China.}%
\thanks{Zihao Zhang is with the Institute of AI for Industries,
Chinese Academy of Sciences, Nanjing, China.}%
\thanks{Xiatian Zhu is with the Surrey Institute for People-Centred
Artificial Intelligence and the Centre for Vision, Speech and
Signal Processing (CVSSP), University of Surrey, Guildford, U.K.}%
\thanks{\textit{Wei Tan and Kunming Luo contributed equally to this work. 
Corresponding author: Lei Xie 
(e-mail: leixie@seu.edu.cn).}}
}
\begin{document}
\maketitle
\input{sec/0_abstract}


\begin{figure*}[!t]
\centering
\includegraphics[width=0.95\textwidth]{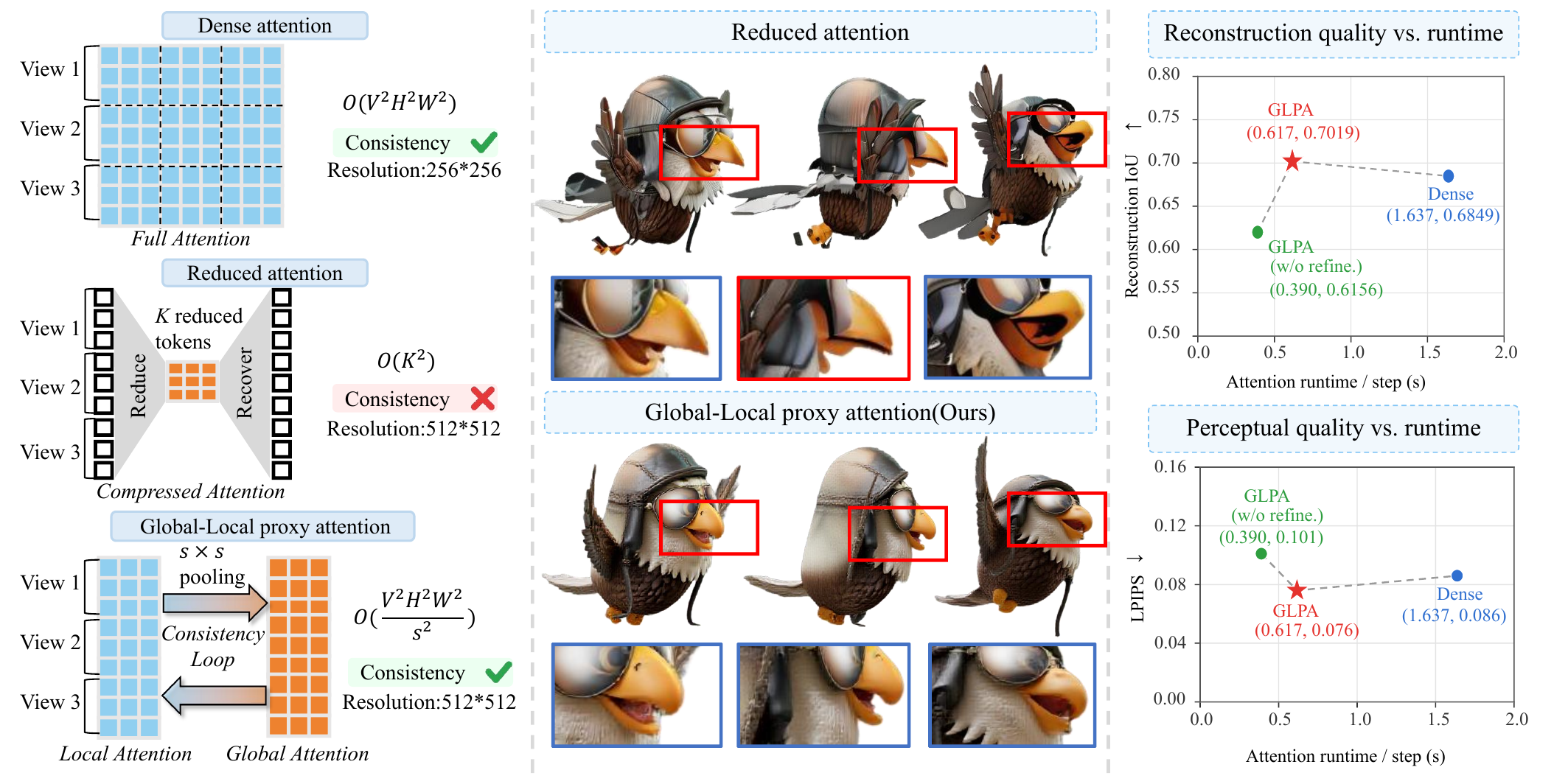}
\caption{Consistent multi-view generation with GLPA. By combining global-local proxy interaction with iterative proxy refinement, GLPA improves cross-view consistency compared with reduced attention while requiring less attention computation than dense attention. }
\label{fig:teaser}
\end{figure*}
\input{sec/1_intro}

\input{sec/2_related}

\input{sec/3_method}

\input{sec/4_experiment}
\input{sec/5_conclusion}
\bibliographystyle{IEEEtran}
\bibliography{main}
\end{document}

%% file: preamble.tex
\usepackage{amssymb}
\usepackage{cite}
\usepackage[hyphens]{url}
\usepackage[dvipsnames]{xcolor}
\usepackage{enumitem}
\usepackage{booktabs}
\usepackage{multirow}
\usepackage{amsmath}
\usepackage{makecell}
\usepackage{marvosym}
\usepackage{tabularx}
\usepackage{bm}
\usepackage{array}
\usepackage{multicol}
\usepackage{enumerate}
\usepackage{float}
\usepackage{placeins}
\usepackage{xspace}
\usepackage{graphicx}
\usepackage{algpseudocode}

\newcounter{glpalgorithm}
\renewcommand{\theglpalgorithm}{\arabic{glpalgorithm}}

%% file: sec/0_abstract.tex
\begin{abstract}

Generating consistent multi-view RGB images and normal maps from a single input image provides complementary appearance and geometric cues for 3D reconstruction. However, maintaining cross-view consistency requires information exchange across viewpoints, and dense cross-view attention becomes computationally expensive as the number of views and spatial resolution increase. In this paper, we propose Global-Local Proxy Attention (GLPA) to reduce cross-view communication costs while preserving fine-grained within-view processing. Specifically, GLPA retains full-resolution interactions within each view and uses compact global proxies for cross-view communication. To further leverage the exchanged information, GLPA incorporates iterative proxy refinement, which rebuilds the proxies from updated features and reuses the same attention parameters for subsequent interactions within a single denoising step. We evaluate GLPA on two structurally different diffusion backbones, FLUX and Stable Diffusion. Experiments on FLUX show that GLPA achieves a $2.65\times$ speedup in attention computation per denoising step over dense attention, while improving reconstruction IoU from 0.6849 to 0.7019 and reducing NVS LPIPS from 0.086 to 0.076. Results on Stable Diffusion further demonstrate the effectiveness of GLPA, with a $1.80\times$ attention speedup and improvements across all five quality metrics.
\end{abstract}

%% file: sec/1_intro.tex
\section{Introduction}

\IEEEPARstart{I}{mmersive} multimedia applications, including virtual and augmented reality and interactive games, rely on 3D assets that can be viewed from different perspectives~\cite{hong2023lrm}. Creating such assets often requires labor-intensive manual modeling or dedicated multi-view capture. Single-image 3D reconstruction offers an alternative by inferring object geometry and appearance from a single input image~\cite{nie2023cpg,qin2026dd3g,huang2025spar3d}. The reconstructed assets should preserve fine geometric and appearance details and maintain consistency across viewpoints~\cite{zhang2026tga}. Achieving these properties efficiently remains an important challenge for single-image 3D content generation.

However, reconstructing a complete 3D object from a single image remains fundamentally ill-posed, as the geometry and appearance of unobserved regions must be inferred from only one observed viewpoint. Recent image-conditioned diffusion models alleviate this ambiguity by transferring strong 2D generative priors to novel-view synthesis~\cite{liu2023zero123}. Building upon this idea, multi-view diffusion models jointly generate images from multiple viewpoints before reconstructing the underlying 3D object~\cite{shi2023mvdream,liu2023syncdreamer}. Wonder3D further couples RGB images with normal maps, providing complementary appearance and geometric cues for downstream reconstruction~\cite{long2023wonder3d}. These approaches have established multi-view generation as an effective intermediate stage in single-image 3D reconstruction, where reconstruction quality strongly depends on both the fidelity of individual views and their consistency across viewpoints.

Maintaining such consistency typically requires explicit information exchange among generated views. Dense cross-view attention allows every generated token to interact with tokens from other viewpoints, but its computational cost grows quadratically with the total number of tokens. As spatial resolution and the number of views increase, this all-to-all interaction quickly becomes a major computational bottleneck. Recent methods therefore seek more efficient cross-view interaction. Era3D replaces dense cross-view attention with row-wise attention under a canonical orthographic camera configuration, substantially reducing the interaction space~\cite{li2024era3d}. However, this efficiency is obtained by imposing structured correspondence and camera constraints on the generated views, which limits the flexibility of cross-view interaction. CTR3D removes this restriction by reducing redundant multi-view tokens before cross-view attention and recovering the full token set afterward, enabling dense and high-resolution multi-view generation without imposing the same row-wise camera constraint~\cite{luo2026ctr3d}. However, its cross-view interaction is performed on the reduced representation, with the original token resolution recovered only after attention. This reduction-and-recovery formulation does not explicitly separate fine-grained local processing from compact cross-view communication. In contrast, these two interactions naturally serve different roles: within-view processing preserves spatial details, whereas cross-view communication mainly conveys information required for consistency across viewpoints.

Motivated by this distinction, we propose Global-Local Proxy Attention (GLPA), which preserves full-resolution within-view interaction while compressing cross-view communication through compact global proxies. As illustrated in Fig.~\ref{fig:teaser}, this design enables consistent multi-view RGB and normal generation for downstream 3D reconstruction while reducing the cost of dense cross-view attention. Specifically, spatially pooled proxy tokens summarize the feature representation of each view and provide a compact pathway for information exchange across viewpoints, while the original full-resolution features remain available for detailed within-view interaction. In this way, cross-view communication is reduced without compressing the feature representation used for fine-grained local processing. A compact proxy representation constructed only once, however, may not fully reflect the feature state after cross-view interaction. The initial proxies are formed before information from other viewpoints has been incorporated into the full-resolution features. After the first interaction, these features contain newly exchanged multi-view information, while the original proxies still summarize the previous state. To further enhance this mechanism, GLPA incorporates iterative proxy refinement, which reconstructs the global proxies from the updated full-resolution features and performs a subsequent interaction using the same attention parameters. This allows the proxy representation to evolve together with the feature state without introducing additional network layers or diffusion sampling steps.

The overall multi-view generation pipeline incorporating GLPA is illustrated in Fig.~\ref{fig:overview}. GLPA combines global-local proxy interaction with bounded iterative proxy refinement within a unified attention design. We instantiate GLPA on a transformer-based FLUX backbone and further adapt the same mechanism to a UNet-based Stable Diffusion backbone to demonstrate its applicability across structurally different diffusion architectures.

Extensive experiments show that a single round of global-local proxy interaction reduces attention runtime, whereas iterative proxy refinement improves generation and reconstruction quality at additional computational cost. On the FLUX backbone, this single-round interaction reduces the attention runtime per denoising step from 1.637 s to 0.390 s compared with dense attention. With iterative refinement, the full GLPA configuration requires 0.617 s per denoising step, achieving a $2.65\times$ attention speedup over dense attention. The generated RGB images and normal maps further support high-quality downstream 3D reconstruction on the GSO benchmark. Results on the SD-based implementation additionally show that the proposed design can be applied to structurally different diffusion backbones.

Our contributions are summarized as follows: (i) we propose GLPA, a unified attention mechanism combining full-resolution within-view interactions, compact proxy-based cross-view communication, and bounded iterative proxy refinement with shared attention parameters; (ii) we implement GLPA on transformer-based FLUX and UNet-based Stable Diffusion to examine its applicability across diffusion architectures; and (iii) we conduct extensive experiments on multi-view generation and 3D reconstruction, demonstrating that GLPA achieves a favorable balance among generation and reconstruction quality, attention runtime, and memory consumption.

%% file: sec/2_related.tex
\begin{figure*}[ht]
\centering
  \includegraphics[width=0.95\textwidth,trim=0 60bp 0 0,
    clip]{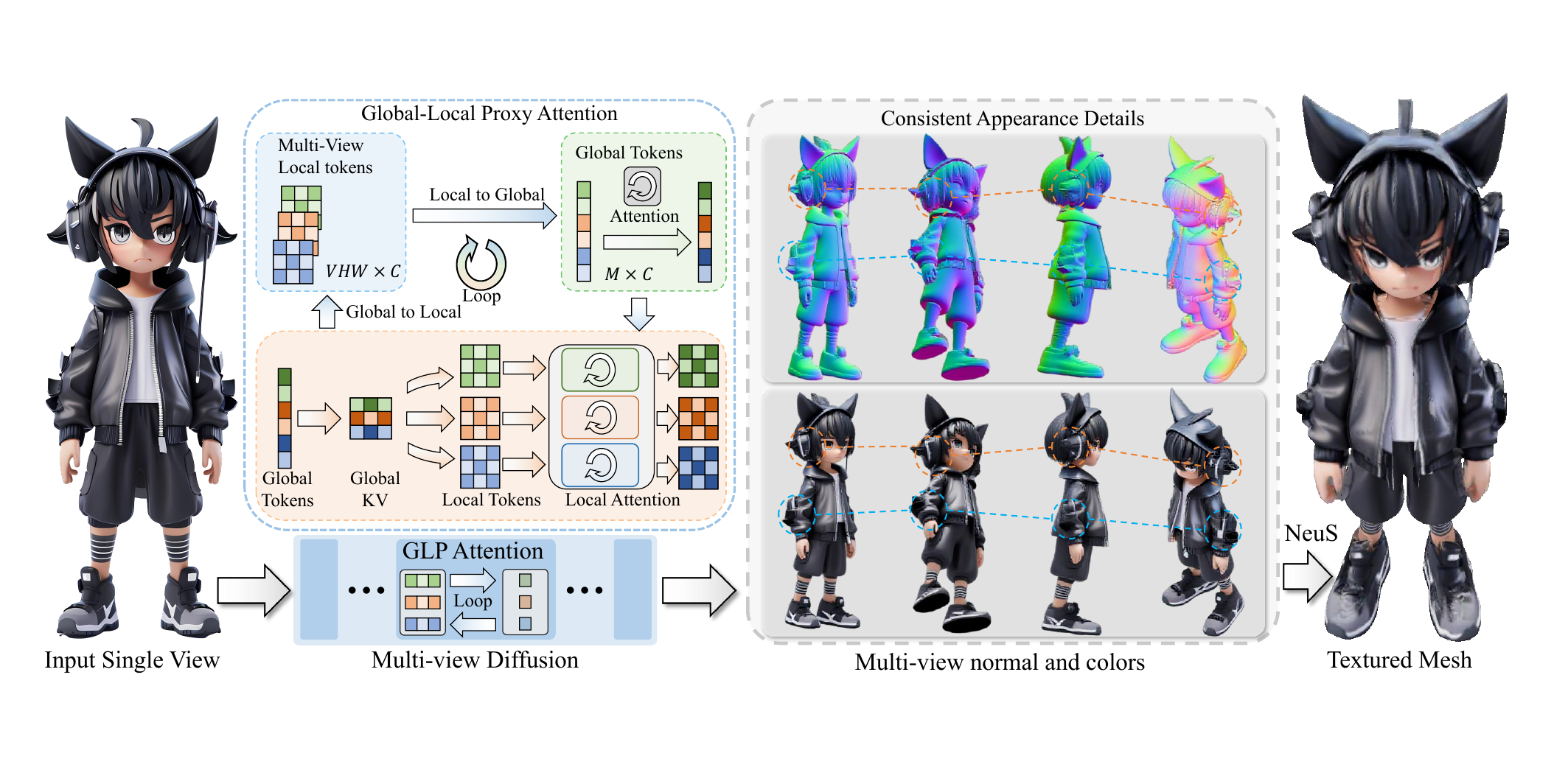}
  \caption{Overall pipeline incorporating GLPA. GLPA combines global-local proxy interaction with iterative proxy refinement to preserve full-resolution within-view interactions and enable efficient cross-view communication. With GLPA integrated into multi-view diffusion, the pipeline generates consistent multi-view RGB images and normal maps from a single input image for downstream textured 3D reconstruction. }
  \label{fig:overview}
\end{figure*}

\section{Related Work}
Recent single-image 3D generation methods can be broadly divided into approaches that directly predict or generate 3D representations and methods that first synthesize consistent multi-view images for subsequent reconstruction. Recent works such as One-2-3-45++~\cite{liu2023one}, LGM~\cite{tang2024lgm}, and CraftsMan3D~\cite{li2024craftsman} have explored different combinations of multi-view generation, feed-forward reconstruction, and 3D-native generative models. Our work follows the multi-view generation paradigm and focuses on improving the efficiency and consistency of cross-view interaction in diffusion-based multi-view synthesis. In this section, we first review recent advances in multi-view and cross-domain generation, and then discuss efficient attention and token reduction strategies that are closely related to our GLPA.

\subsection{Multi-view diffusion}
Building on the generative priors of pretrained 2D diffusion models, novel-view synthesis has expanded from camera-conditioned single-view prediction to joint multi-view generation~\cite{liu2023zero123,shi2023mvdream,wang2023imagedream,shi2023zero123++,tang2024mvdiffusion++}. Cross-view feature interaction is widely adopted during joint denoising to maintain structural and appearance consistency across generated views. Spatial geometry also provides useful cues for cross-view communication. Existing methods employ pixel correspondences and epipolar constraints to guide feature interaction~\cite{tang2023MVDiffusion,kant2024spad}, or associate information across viewpoints through 3D-aware feature aggregation and depth-based reprojection~\cite{liu2023syncdreamer,hu2024mvdfusion,cao2025mvgenmaster}. Geometric correspondences have also been used to supervise attention maps for improved spatial consistency~\cite{kwon2026cameo}.
For downstream 3D reconstruction, generated multi-view observations need to capture both appearance and geometry consistently. Recent approaches combine RGB synthesis with surface normal or depth prediction to provide complementary geometric cues for reconstruction~\cite{long2023wonder3d,guizilini2025mvgd}. Video diffusion priors have also been employed to synthesize object views along camera trajectories~\cite{chen2025v3d,voleti2025sv3d,yang2024hi3d}. As the number of views and spatial resolution increase, cross-view communication becomes computationally expensive, motivating more efficient approaches to high-resolution and dense multi-view generation~\cite{li2024era3d,luo2026ctr3d}.
\subsection{Efficient cross-view interaction}
Reducing redundant token computations is a common strategy for improving attention efficiency in Transformers~\cite{kong2026tokenreduction}. Token merging shortens the sequence involved in attention by combining similar features~\cite{bolya2023tome}, and has been applied to Stable Diffusion-based image generation and zero-shot video editing~\cite{bolya2023tomesd,li2024vidtome}. Compressing key and value representations while retaining the original queries can also reduce attention costs~\cite{smith2024todo,chen2024pixart}, while EDiT combines spatial key-value compression with linear attention for efficient diffusion generation~\cite{Becker2025EDiT}. Attention can also operate at different spatial granularities, combining fine-grained local interactions with coarser long-range information aggregation~\cite{yang2021focalattention,pan2023hilo}. In multi-view diffusion, adjustments to the interaction range and feature representations are further reflected in the design of cross-view communication. Era3D employs a canonical orthographic camera configuration to restrict cross-view attention to corresponding image rows, whereas CTR3D uses token reduction and recovery to perform cross-view interactions on compressed multi-view representations~\cite{li2024era3d,luo2026ctr3d}.
Global information exchange can also be mediated by a small set of intermediate tokens, reducing dense interactions among the original features. Agent Attention and DAM employ intermediate representations to reduce the interaction scale between queries and keys/values, with DAM introducing dynamic attention mediators into diffusion Transformers~\cite{han2024agentattention,pu2024dam}. PT-DiT constructs proxies from features aggregated over spatial or spatiotemporal windows, establishes global dependencies through attention among proxies, and employs window and shifted-window attention to process local features. This design has been applied to text-conditioned image, video, and multi-view generation~\cite{wang2025pttiv}. For single-image-conditioned joint multi-view RGB/normal generation, a key question is how to reduce cross-view communication costs while preserving fine-grained within-view interactions. GLPA addresses this through global-local proxy interaction, which retains full-resolution within-view attention and uses compact global proxies for cross-view communication.
\begin{figure*}[t]
    \centering
    \includegraphics[width=0.75\textwidth]{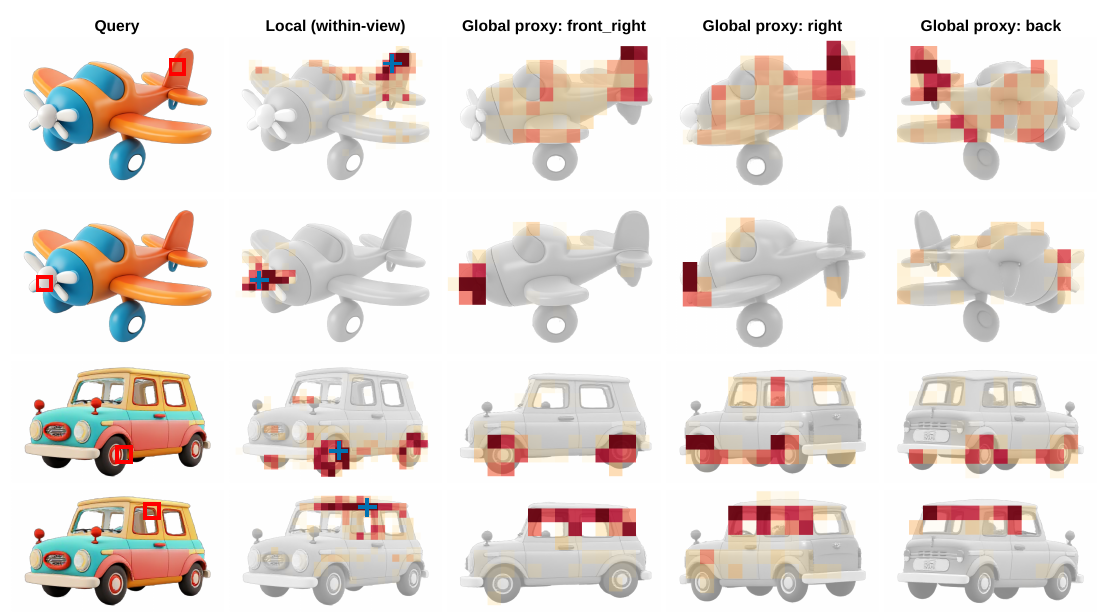}
    \caption{Visualization of a single round of global-local proxy interaction in GLPA. Each row shows a query cell on the front view (red box), the within-view attention at this location, and the global attention of this cell over the proxies of three other views, displayed at their associated spatial locations. The proxy pathway reaches the corresponding parts across viewpoints, while the local pathway preserves fine within-view structure.}
    \label{fig:proxy_attention}
\end{figure*}
\subsection{Looped Transformers}
Shared-parameter computation allows Transformer modules to be applied repeatedly to updated features without requiring a complete set of layer parameters for each iteration~\cite{dehghani2019universaltransformers}. Depth-specific low-rank adaptation and loop-level hyper-connections provide additional flexibility in feature transformation and state propagation when Transformer modules are reused~\cite{bae2025relaxedrecursive,zeitoun2026hyperloop}. Hyper-SET develops a recurrent-depth Transformer based on hyperspherical energy minimization, where feature states are updated through iterative optimization~\cite{hu2026hyperset}. Perceiver repeatedly reads the input to update compact latent representations, allowing subsequent attention operations to build on previously aggregated information~\cite{jaegle2021perceiver}. For visual generation, ELT reuses Transformer modules across iterations for image and video synthesis, using intra-loop self-distillation to supervise intermediate predictions~\cite{goyal2026elt}. For proxy-based communication in multi-view diffusion, a relevant question is how features updated by an initial interaction can contribute to subsequent cross-view information exchange. Within a single diffusion denoising step, GLPA updates the full-resolution working state through residual feedback with a bounded coefficient, reconstructs global proxies from the updated features, and performs subsequent interactions with the same attention layer, sharing its parameters across interaction rounds.

%% file: sec/3_method.tex
\section{Method}

We integrate GLPA into a multi-view diffusion model for high-quality multi-view generation from a single input image, as illustrated in Fig.~\ref{fig:overview}. Given an input image, the diffusion model generates multi-view RGB images and normal maps, which are then used to reconstruct a textured 3D mesh with an existing method such as NeuS~\cite{wang2021neus}. GLPA reduces dense cross-view computation by preserving full-resolution within-view interaction and compressing cross-view communication through global proxies.

Specifically, GLPA consists of two main components: global-local proxy interaction (Sec.~\ref{algo:method_2}) and iterative proxy refinement (Sec.~\ref{algo:method_3}), as outlined in Alg.~\ref{alg:glp_refinement}. During a global-local proxy interaction, each view preserves full-resolution local attention, while pooled proxy tokens summarize information from different views and enable efficient cross-view communication. Based on the updated features, the global proxies are further refined in the subsequent interaction round, allowing cross-view information to be progressively updated and propagated. 
\subsection{Multi-view latent generation}\label{algo:method_1}
Given an input image $I^{in}$, we jointly generate color and normal maps $\{(I_v, N_v)\}_{v=1}^{V}$ at $V$ predefined viewpoints. Color images record appearance while normal maps characterize surface geometry, and together they constrain the complete shape of the same object. Following existing multi-view diffusion frameworks, generation is performed in the latent space of the diffusion model and decoded into image space to obtain the final color and normal maps.
At a given attention layer, let $U \in \mathbb{R}^{S \times C}$ denote the normalized input of the attention layer, where $S$ is the total number of generated tokens across all views and domains. Directly performing dense cross-view attention requires computing pairwise attention scores among all generated tokens, resulting in $O(S^2)$ complexity. This quadratic growth becomes the main computational bottleneck when increasing the number of views or spatial resolution. Our design preserves full-resolution within-view interaction and compresses only the representations responsible for cross-view communication.
For interaction round $r \in \{1,2\}$, let $X^{(r)} \in \mathbb{R}^{S \times C}$ denote the full-resolution working state, with $X^{(1)} = U$; the second-round state $X^{(2)}$ is defined in Sec.~\ref{algo:method_3}. Within $X^{(r)}$, the tokens of view $v$ and domain $d \in \{\mathrm{RGB}, \mathrm{normal}\}$ form a feature grid $X^{(r)}_{v,d} \in \mathbb{R}^{H \times W \times C}$ with spatial tokens $x^{(r)}_{v,d,i} \in \mathbb{R}^{C}$, so $S = 2VHW$ for equal-sized grids. We use $x^{(r)}_i$ when the view and domain indices are implicit.
\subsection{Global-local proxy interaction}\label{algo:method_2}
Our design assigns different information roles to cross-view and within-view interaction. Cross-view communication exchanges information shared across viewpoints, while within-view interaction processes spatially detailed features such as silhouettes, normal discontinuities, and textures. We therefore retain full-resolution within-view processing and use compact spatial summaries for cross-view communication.

\subsubsection{Local interactions} Generated tokens are organized into local groups by viewpoint, each group containing the RGB and normal tokens of a single view. Tokens within a group interact at their original resolution, and no direct full-resolution connection is formed across views. The proxy pathway does not spatially compress or replace the full-resolution within-view token path.

\subsubsection{Global proxies} Cross-view communication is carried by a compact set of global proxies. For each view, we apply channel-wise spatial average pooling with kernel size $s\times s$ and stride $s$, separately within the color and normal grids. The grids are divisible by $s$, so the pooling regions do not overlap or cross domain boundaries, and pooling preserves the channel dimension. Average pooling provides deterministic spatial aggregation within each view/domain partition without a learned token-selection module.
For each view $v$ and domain $d$, let $j$ denote the pooling-cell index and $\Omega_j$ the set of spatial token indices within that cell. The pooled proxy feature in round $r$ is computed as
\begin{equation}
p^{(r)}_{v,d,j}
=
\frac{1}{|\Omega_j|}
\sum_{i\in\Omega_j} x^{(r)}_{v,d,i}.
\end{equation}
The proxies from all views and both domains are concatenated into the pooled proxy bank $P^{(r)} = \mathrm{Pool}(X^{(r)}) \in \mathbb{R}^{M \times C}$, yielding $M=\frac{S}{s^2}$ pooled tokens in total. Since each pooled proxy only summarizes its own view and domain, we further apply self-attention over the entire proxy bank so that proxies from different views and domains can exchange information:
\begin{equation}
\begin{aligned}
\widetilde{P}^{(r)}
&= \mathrm{ProxySA}_{\theta}(P^{(r)}) \\
&= \mathrm{Softmax}\left(
\frac{Q(P^{(r)})\,K(P^{(r)})^{\top}}{\sqrt{d_h}}
\right)V(P^{(r)}),
\end{aligned}
\end{equation}
where $\theta$ denotes the attention parameters of the current layer, $Q(\cdot)$, $K(\cdot)$, and $V(\cdot)$ are its query, key, and value projections, and $d_h$ is the per-head channel dimension. Proxy self-attention reuses the same projections as the cross-view aggregation below and introduces no additional parameters. We refer to $P^{(r)}$ as the pooled proxies and to $\widetilde{P}^{(r)}$ as the contextualized proxies.

\subsubsection{Cross-view aggregation} For a full-resolution query token $x^{(r)}_i$, let $\mathcal{L}^{(r)}_i$ denote its full-resolution local token group in $X^{(r)}$. The keys and values available to this query are constructed by concatenating the local full-resolution tokens and the contextualized global proxies:
\begin{equation}
\begin{aligned}
K^{(r)}_i &= \big[K(\mathcal{L}^{(r)}_i);\, K(\widetilde{P}^{(r)})\big], \\
V^{(r)}_i &= \big[V(\mathcal{L}^{(r)}_i);\, V(\widetilde{P}^{(r)})\big].
\end{aligned}
\end{equation}
The resulting attention is computed as:
\begin{equation}
\mathrm{Attn}(x^{(r)}_i)
=
\mathrm{Softmax}\left(
\frac{Q(x^{(r)}_i)\,(K^{(r)}_i)^{\top}}{\sqrt{d_h}}
\right)V^{(r)}_i.
\end{equation}
Cross-view information is accessed through $\widetilde{P}^{(r)}$, while $\mathcal{L}^{(r)}_i$ retains full-resolution local interaction. As illustrated in Fig.~\ref{fig:proxy_attention}, a query in one view
attends to proxy tokens associated with spatial regions in other views. We denote the attention output over all full-resolution queries in round $r$ as
\begin{equation}
\Delta^{(r)} = \mathrm{GLPAttn}_{\theta}(X^{(r)}, \widetilde{P}^{(r)}) \in \mathbb{R}^{S \times C}.
\end{equation}
Cross-view readout through the global proxies requires $SM$ pairwise interactions, compared with $S^2$ in dense attention. Proxy self-attention adds a further $M^2$ interactions.

\begin{figure}[!t]
\small
\refstepcounter{glpalgorithm}
\label{alg:glp_refinement}
\hrule
\vspace{0.4ex}
\noindent
\textbf{Algorithm \theglpalgorithm}\enspace
Global-Local Proxy Attention
\par
\vspace{0.4ex}
\hrule
\vspace{0.6ex}
\begin{algorithmic}[1]
\State \textbf{Input:} Normalized features
    $U\in\mathbb{R}^{S\times C}$,
    stride $s$, parameters $\theta,\alpha,\gamma$.
\State \textbf{Output:} Refined attention output $\Delta^{(2)}$.

\State \textbf{Stage 1: First-round Interaction}
\State $X^{(1)} \gets U$
\State $P^{(1)} \gets \operatorname{Pool}_{s}(X^{(1)})$
\State $\widetilde{P}^{(1)} \gets
    \operatorname{ProxySA}_{\theta}(P^{(1)})$
\State $\Delta^{(1)} \gets
    \Call{Interaction}{X^{(1)},\widetilde{P}^{(1)},\theta}$

\State \textbf{Stage 2: Bounded Feedback \& Proxy Rebuild}
\State $X^{(2)} \gets
    U+\gamma\tanh(\alpha)\,\Delta^{(1)}$
\State $P^{(2)} \gets \operatorname{Pool}_{s}(X^{(2)})$
\Statex \hspace{\algorithmicindent}
    \(\triangleright\) Rebuild from updated features
\State $\widetilde{P}^{(2)} \gets
    \operatorname{ProxySA}_{\theta}(P^{(2)})$
\State $\Delta^{(2)} \gets
    \Call{Interaction}{X^{(2)},\widetilde{P}^{(2)},\theta}$
\State \Return $\Delta^{(2)}$

\smallskip
\Function{Interaction}{$X^{(r)},\widetilde{P}^{(r)},\theta$}
    \For{$i=1,\ldots,S$}
        \State $\mathcal{L}_i^{(r)} \gets$
            same-view RGB/normal tokens from $X^{(r)}$
        \State $K_i^{(r)} \gets
            [K(\mathcal{L}_i^{(r)});K(\widetilde{P}^{(r)})]$
        \State $V_i^{(r)} \gets
            [V(\mathcal{L}_i^{(r)});V(\widetilde{P}^{(r)})]$
        \State $a_i^{(r)} \gets
            \operatorname{Softmax}\!\left(
            \dfrac{Q(x_i^{(r)})(K_i^{(r)})^\top}
                  {\sqrt{d_h}}
            \right)V_i^{(r)}$
    \EndFor
    \State $\Delta^{(r)} \gets
        \Call{OutputProjection}
        {\{a_i^{(r)}\}_{i=1}^{S},\theta}$
    \State \Return $\Delta^{(r)}$
\EndFunction
\end{algorithmic}
\vspace{0.4ex}
\hrule
\end{figure}

\subsection{Iterative proxy refinement}\label{algo:method_3}
In the single-round interaction described above, global proxies are constructed from the input features of the current layer. After the first interaction, the updated tokens already contain information propagated from other views. We therefore rebuild the global proxies from these updated features and perform a second interaction, allowing the cross-view representation to be further refined based on the newly exchanged information. 

\subsubsection{Bounded feedback} In the first round, $X^{(1)} = U$, and the contextualized proxies and attention output are
\begin{equation}
\begin{aligned}
\widetilde{P}^{(1)} &= \mathrm{ProxySA}_{\theta}\big(\mathrm{Pool}(X^{(1)})\big), \\
\Delta^{(1)} &= \mathrm{GLPAttn}_{\theta}(X^{(1)}, \widetilde{P}^{(1)}).
\end{aligned}
\end{equation}
The working state of the second round is
\begin{equation}
X^{(2)} = U + g\,\Delta^{(1)}, 
\qquad
g = \gamma \tanh(\alpha),
\end{equation}
where $\alpha$ is a learned scalar local to the attention layer and $\gamma > 0$ is a fixed hyperparameter. Hence $|g| \leq \gamma$: the coefficient of the residual update is bounded. The second-round state is a gated residual update anchored to $U$.

\subsubsection{Proxy reconstruction} In the second round, the proxies are re-pooled from $X^{(2)}$ and contextualized again by proxy self-attention, and the queries, keys, and values are recomputed from $X^{(2)}$ following the construction in Sec.~\ref{algo:method_2}:
\begin{equation}
\begin{aligned}
P^{(2)} &= \mathrm{Pool}(X^{(2)}), \\
\widetilde{P}^{(2)} &= \mathrm{ProxySA}_{\theta}(P^{(2)}), \\
\Delta^{(2)} &= \mathrm{GLPAttn}_{\theta}(X^{(2)}, \widetilde{P}^{(2)}).
\end{aligned}
\end{equation}
Reusing the first-round proxies would keep the cross-view representation tied to the previous feature state, whereas reconstructing the proxies allows the updated full-resolution features to participate in the subsequent cross-view interaction. The two rounds share the same $Q/K/V$ projections, attention-normalization parameters, and output projection where applicable, without introducing round-specific attention weights. Gradients are propagated through the feedback path, and each refined attention layer introduces only one learnable scalar $\alpha$, while the fixed scale $\gamma$ is shared as a hyperparameter. The pooling operation itself remains parameter-free. Refinement is applied only to attention layers that use global-local proxy interaction. The entire refinement process is performed within a single denoising step by reusing the same attention layer, without adding extra sampling steps or network layers. Intermediate outputs provide feedback to the subsequent round, and the output of the final executed round is used as the attention result.

\begin{figure*}[t]
    \centering
    \includegraphics[width=0.85\textwidth]{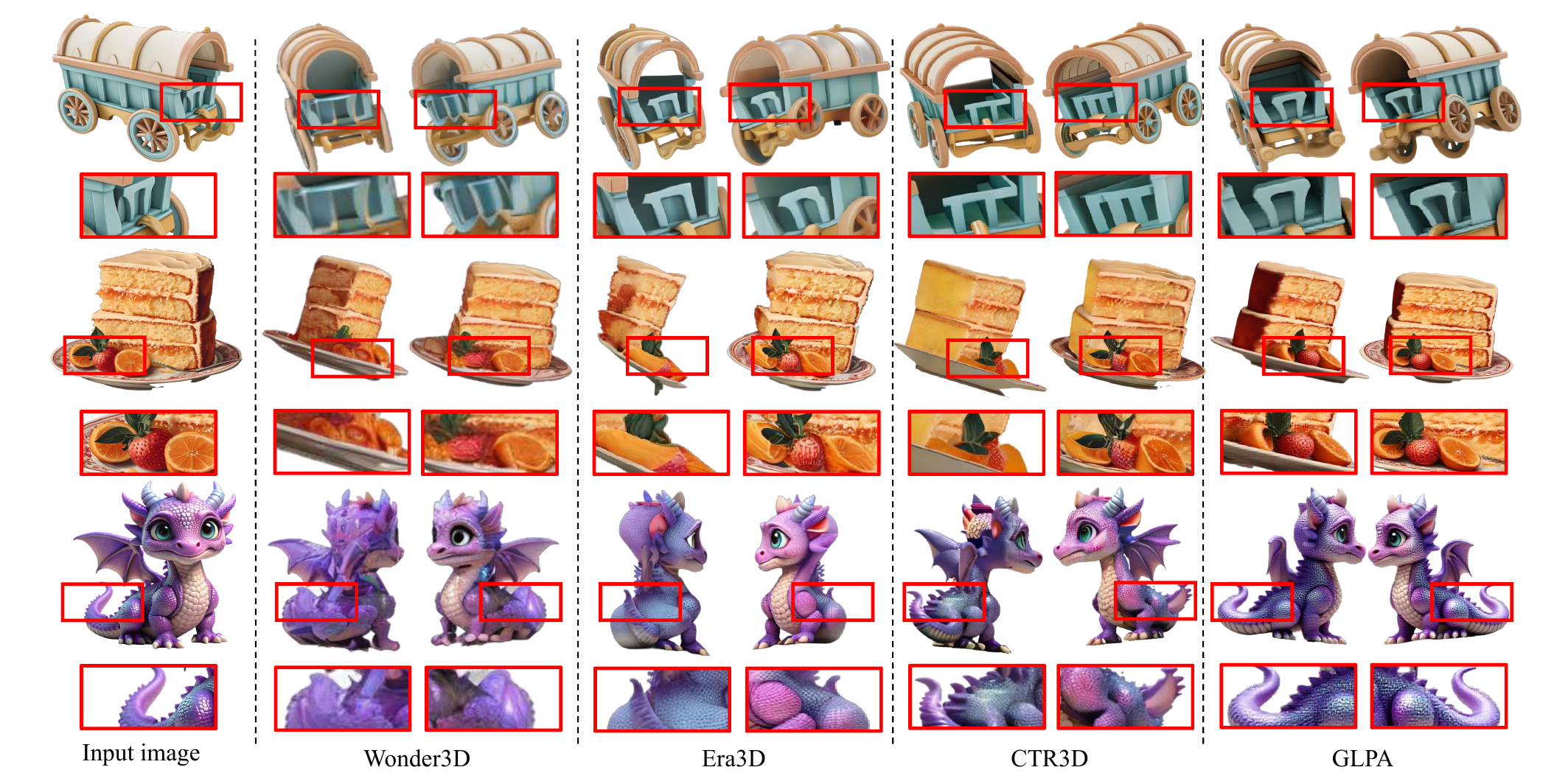}
    \caption{Qualitative comparison with baseline methods on novel view synthesis.}
    \label{fig:figure4}
\end{figure*}

\subsection{Instantiation on a UNet-Based Backbone}\label{algo:method_4}
On the SD UNet backbone, GLPA retains the same proxy construction and iterative refinement as the FLUX implementation. Local attention groups follow the UNet's factorized view and domain structure, with each group containing tokens from a single view and domain. Global proxies aggregate information across views and domains and incorporate the corresponding view and domain embeddings. The SD attention projections and readout parameters are shared across refinement rounds.



%% file: sec/4_experiment.tex
\section{Experiments}


\begin{table}[t]
  \caption{Quantitative Comparison on the GSO Benchmark Using Chamfer Distance (CD) and IoU for 3D Reconstruction, and LPIPS, SSIM, and PSNR for Novel View Synthesis.}
  \label{tab:main}
  \centering
  \footnotesize
  \renewcommand{\arraystretch}{1.15}
  \setlength{\tabcolsep}{3pt}
  \begin{tabular*}{\columnwidth}{@{\extracolsep{\fill}}l|ccccc@{}}
    \hline
    Methods & CD $\downarrow$ & IoU $\uparrow$ &
    LPIPS $\downarrow$ & SSIM $\uparrow$ & PSNR $\uparrow$ \\
    \hline

    RealFusion~\cite{melaskyriazi2023realfusion}
      & 0.0819 & 0.2714 & 0.283 & 0.722 & 15.26 \\
    Zero-1-to-3~\cite{liu2023zero123}
      & 0.0339 & 0.5035 & 0.166 & 0.779 & 18.93 \\
    One-2-3-45~\cite{one2345}
      & 0.0629 & 0.4086 & - & - & - \\
    Shap-E~\cite{jun2023shape}
      & 0.0436 & 0.3584 & - & - & - \\
    Magic123~\cite{qian2023magic123}
      & 0.0516 & 0.4528 & - & - & - \\
    SyncDreamer~\cite{liu2023syncdreamer}
      & 0.0261 & 0.5421 & 0.146 & 0.798 & 20.05 \\
    Wonder3D~\cite{long2023wonder3d}
      & 0.0248 & 0.5678 & 0.141 & 0.811 & 20.83 \\
    LGM~\cite{tang2024lgm}
      & 0.0259 & 0.5628 & - & - & - \\
    Era3D~\cite{li2024era3d}
      & 0.0217 & 0.5973 & 0.126 & 0.837 & 22.74 \\
    CTR3D~\cite{luo2026ctr3d}
      & 0.0206 & 0.6447 & 0.081 & 0.914 & 23.47 \\

    Dense (SD)
      & 0.0335 & 0.5505 & 0.119 & 0.904 & 21.89 \\
    GLPA (SD)
      & 0.0224 & 0.6641 & 0.098 & 0.911 & 23.72 \\

    Dense (FLUX)
      & 0.0193 & 0.6849 & 0.086 & 0.922
      & \textbf{28.03} \\
    GLPA (FLUX)
      & \textbf{0.0185} & \textbf{0.7019}
      & \textbf{0.076} & \textbf{0.927} & 27.80 \\

    \hline
  \end{tabular*}
\end{table}

\begin{figure}[t]
\centering
  \includegraphics[width=0.48\textwidth]{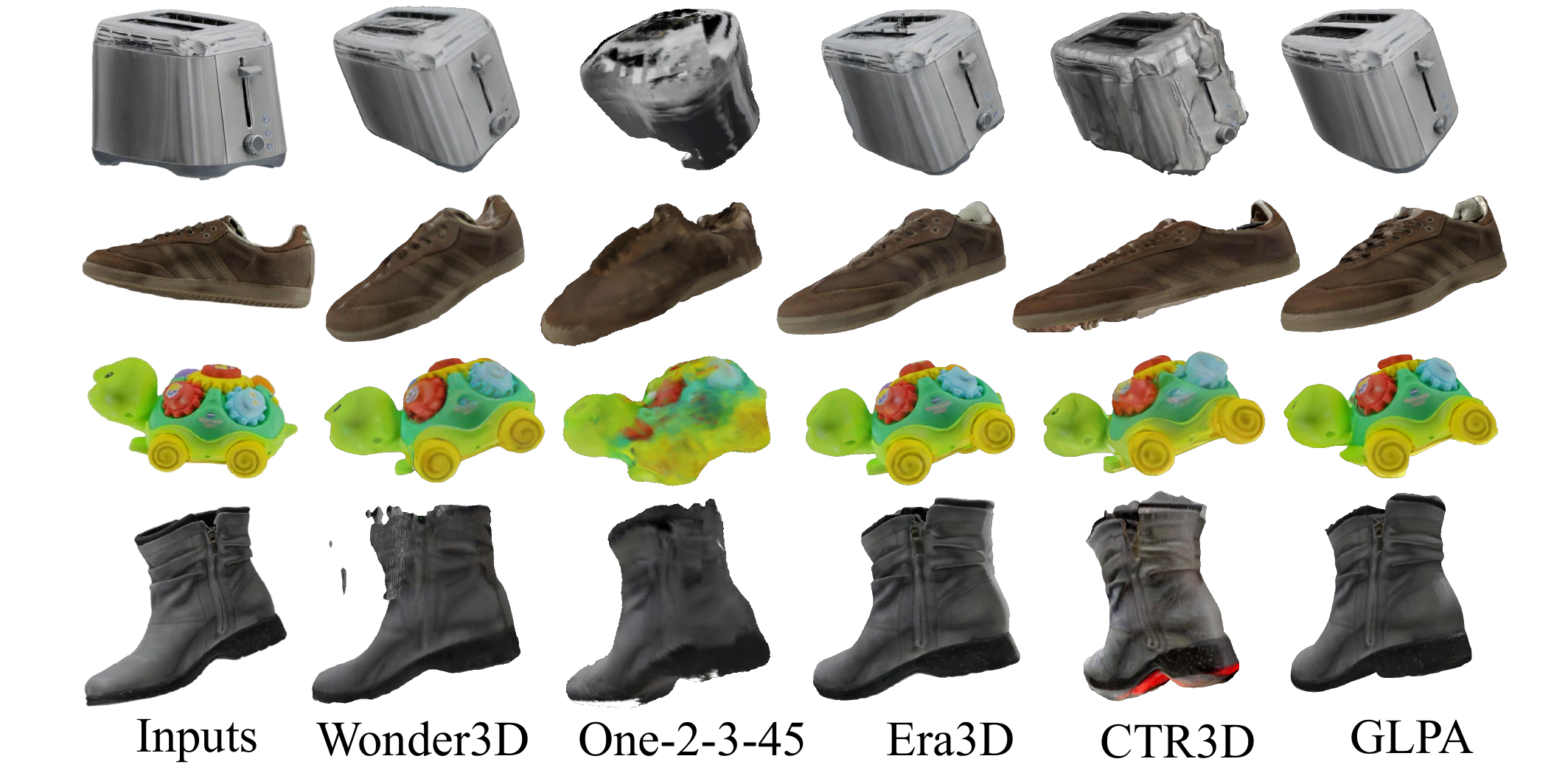}
  \caption{Qualitative comparison of 3D reconstruction results on the GSO dataset. The GLPA-based pipeline generates the best results with more views and high resolution, compared with existing methods.}
  \label{fig:gso_mesh}
\end{figure}

\subsection{Datasets}
Following previous methods~\cite{li2024era3d,long2023wonder3d,luo2026ctr3d}, we train the diffusion models equipped with GLPA on the LVIS subset of Objaverse~\cite{deitke2023objaverse}, which consists of approximately 32,000 high-quality 3D objects. For multi-view data construction, each object is normalized to a unit scale and placed at the scene center. We then rendered RGB images and corresponding normal maps at a resolution of \(512\times512\) from eight predefined viewpoints. Specifically, we use six horizontal views, including front, back, left, right, front-right, and front-left, together with a front-right-top view and a front-left-down view. Random rotations are further applied to the 3D objects during rendering to increase the diversity of training samples. Following previous works, we evaluate the GLPA-based pipeline on the Google Scanned Objects (GSO) dataset~\cite{gso}, a commonly used benchmark for multi-view generation and 3D reconstruction. We further evaluate our method on images collected from the Internet to examine its generalization ability in real-world scenarios. 

\subsection{Metrics}
Our experiments cover novel view synthesis (NVS) and 3D reconstruction. NVS quality is measured with Learned Perceptual Image Patch Similarity (LPIPS)~\cite{zhang2018lpips}, which captures perceptual similarity between synthesized and reference views while tolerating slight spatial misalignment. For 3D reconstruction, Chamfer Distance (CD) and Volume Intersection over Union (IoU) are employed to measure the geometric agreement between reconstructed meshes and ground-truth shapes.

\begin{figure*}[!t]
  \centering
  \includegraphics[width=0.85\textwidth]{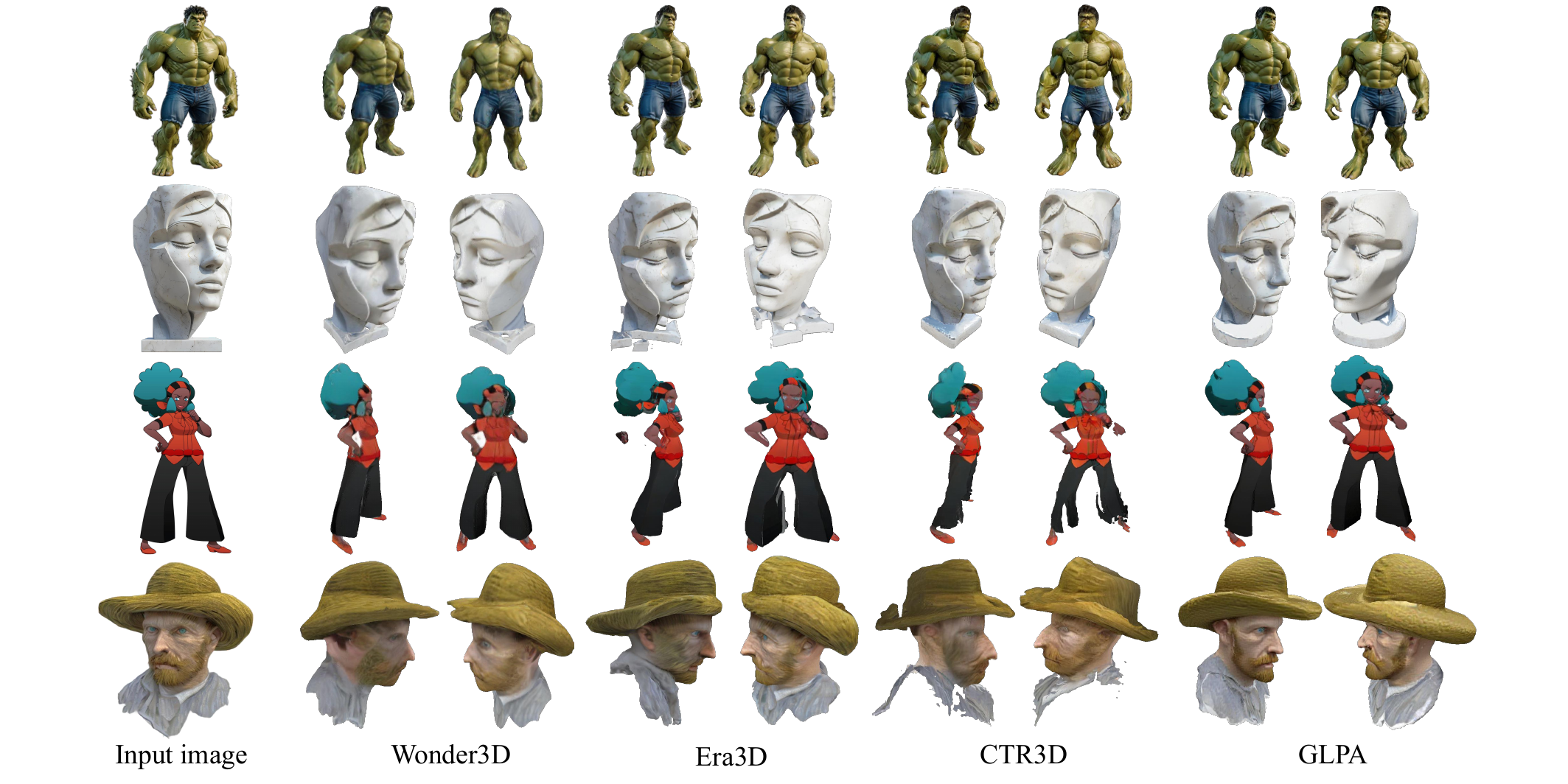}
  \caption{Qualitative comparison of 3D reconstructions on in-the-wild data. Compared with previous methods, the GLPA-based pipeline produces more complete and higher-quality 3D models.}
  \label{fig:wild mesh}
\end{figure*}

\begin{table*}[!tp]
  \caption{Attention Runtime and Quality Comparison of Dense Attention and GLPA Variants on FLUX and SD Backbones.}
  \label{tab:flux_components}
  \centering
  \footnotesize
  \setlength{\tabcolsep}{4pt}
  \renewcommand{\arraystretch}{1.2}
  \begin{tabular}{@{}l|cccccccc@{}}
    \hline
    Methods & N-view & Resolution &
    \makecell{Attention\\runtime/step (s)} &
    CD $\downarrow$ & IoU $\uparrow$ &
    LPIPS $\downarrow$ & SSIM $\uparrow$ & PSNR $\uparrow$ \\
    \hline

    \multicolumn{9}{@{}l}{\textit{FLUX.1-dev Backbone}} \\
    Dense & 8 & 512 & 1.637 & 0.0193 & 0.6849 & 0.086 & 0.922 & \textbf{28.03} \\
    GLPA (w/o refinement) & 8 & 512 & \textbf{0.390} & 0.0246 & 0.6156 & 0.101 & 0.914 & 26.75 \\
    GLPA (reuse proxy) & 8 & 512 & 0.598 & 0.0254 & 0.6246 & 0.087 & 0.920 & 27.44 \\
    GLPA & 8 & 512 & 0.617 & \textbf{0.0185} & \textbf{0.7019} & \textbf{0.076} & \textbf{0.927} & 27.80 \\
    \hline

    \multicolumn{9}{@{}l}{\textit{SD2.1-UnCLIP Backbone}} \\
    Dense & 12 & 512 & 0.730 & 0.0335 & 0.5505 & 0.119 & 0.904 & 21.89 \\
    GLPA (w/o refinement) & 12 & 512 & \textbf{0.277} & 0.0246 & 0.6432 & 0.133 & 0.892 & 21.14 \\
    GLPA (reuse proxy) & 12 & 512 & 0.399 & 0.0252 & 0.6374 & 0.125 & 0.898 & 21.60 \\
    GLPA & 12 & 512 & 0.405 & \textbf{0.0224} & \textbf{0.6641} & \textbf{0.098} & \textbf{0.911} & \textbf{23.72} \\
    \hline
  \end{tabular}
\end{table*}

\subsection{Implementation details}
We implement GLPA on two diffusion backbones, SD2.1-UnCLIP~\cite{rombach2022high} and FLUX.1-dev. For the SD2.1 configuration, we generate 12 RGB and normal views. The model is trained on 8 A100 GPUs for 40,000 training steps with a total batch size of 128. We initialize the learning rate at $1\times10^{-4}$ and decrease it to $5\times10^{-5}$ after 5,000 training steps. During inference, we use DDIM~\cite{ddim} with 40 denoising steps and a guidance scale of 3.0. For the FLUX.1-dev configuration, we generate 8 RGB and normal views. We train the model on 8 A100 GPUs for 10,000 steps with a total batch size of 32. The optimization is performed using Prodigy with a base learning rate of 1.0. For multi-view generation with FLUX.1-dev, we adopt flow-matching-based Euler sampling with 28 denoising steps and set the guidance scale to 3.5. After multi-view generation, we follow previous methods and reconstruct the 3D meshes using NeuS~\cite{wang2021neus}, followed by a texture refinement step. 
\subsection{Experimental Results}



\subsubsection{Novel view synthesis} We first evaluate the novel view synthesis performance on the GSO dataset using three commonly adopted metrics, namely LPIPS, SSIM, and PSNR. As shown in Table~\ref{tab:main}, on the FLUX backbone, GLPA improves LPIPS and SSIM over dense attention, with a slight decrease in PSNR. GLPA (FLUX) also outperforms CTR3D across all three NVS metrics. On the SD backbone, GLPA further improves all three metrics over the corresponding dense-attention baseline, demonstrating its effectiveness across different diffusion backbones. We further provide qualitative comparisons in Fig.~\ref{fig:figure4} to examine the visual quality of the generated results. Compared with the recent state-of-the-art method CTR3D, our approach produces novel views with better visual fidelity on regular cases while preserving fine-grained object details and cross-view consistency (as in row $1$). CTR3D performs cross-view interaction on reduced tokens before recovering dense features. Although efficient, this reduction-and-recovery process may weaken fine-grained details and cross-view appearance consistency in some regions. In contrast, GLPA preserves the full-resolution within-view path and compresses only cross-view communication, using pooled global proxies for efficient information exchange while retaining detailed local features. This Global-Local design leads to more coherent multi-view generation, especially for repeated parts and regions that require consistent cross-view correspondence.


\begin{table*}[!tp]
  \caption{Effect of Pooling Stride and Interaction Rounds on Attention Efficiency, Novel View Synthesis, and 3D Reconstruction.}
  \label{tab:settings_ablation}
  \centering
  \footnotesize
  \renewcommand{\arraystretch}{1.2}
  \setlength{\tabcolsep}{6pt}
  \begin{tabular}{@{}cc|cccccc@{}}
    \hline
    $s$ & $R$ & \makecell{Attention\\runtime/step (s)} & CD $\downarrow$ & IoU $\uparrow$ & LPIPS $\downarrow$ & SSIM $\uparrow$ & PSNR $\uparrow$ \\
    \hline
    2 & 1 & 0.390 & 0.0246 & 0.6156 & 0.101 & 0.914 & 26.75 \\
    2 & 2 & 0.617 & \textbf{0.0185} & \textbf{0.7019} & \textbf{0.076} & \textbf{0.927} & \textbf{27.80} \\
    2 & 3 & 0.799 & 0.0262 & 0.6210 & 0.084 & 0.921 & 27.41 \\
    4 & 2 & 0.370 & 0.0308 & 0.5578 & 0.096 & 0.917 & 27.47 \\
    8 & 2 & \textbf{0.324} & 0.0404 & 0.4496 & 0.125 & 0.908 & 25.75 \\
    \hline
  \end{tabular}
\end{table*}

\subsubsection{Reconstruction} To further evaluate our method on the 3D reconstruction task, we present the quantitative reconstruction results on the GSO dataset in Table~\ref{tab:main}. On the FLUX backbone, GLPA improves both CD and IoU over dense attention, achieving a CD of 0.0185 and an IoU of 0.7019. It also outperforms CTR3D (0.0206 and 0.6447, respectively), indicating improved geometric reconstruction accuracy. On the SD backbone, GLPA similarly improves both metrics over the corresponding dense-attention baseline. We also provide qualitative comparisons of the reconstructed meshes with previous methods in Fig.~\ref{fig:gso_mesh}. We also validate our method on in-the-wild images in Fig.~\ref{fig:wild mesh}. The results (as in row $1$ and $4$) indicate that inconsistent multi-view observations are insufficient to recover complete object structures, leading to lower-quality reconstructions. In contrast, our method generates more consistent views across different viewpoints, allowing us to produce complete and high-quality 3D models.

\subsection{Ablation Study}

\subsubsection{Iterative Refinement}
To evaluate global-local proxy interaction and iterative proxy refinement on both FLUX and SD backbones, we compare dense attention, GLPA without refinement, GLPA with the first-round proxies reused in the second round, and the full model. As shown in Table~\ref{tab:flux_components}, GLPA without refinement substantially reduces attention runtime but yields lower generation and reconstruction quality than dense attention on the FLUX backbone. Iterative refinement improves performance on both tasks, enabling the full model to achieve better reconstruction accuracy and competitive novel view synthesis quality relative to dense attention. Compared with the proxy-reuse variant, the full model reduces CD from 0.0254 to 0.0185 and increases IoU from 0.6246 to 0.7019, while also improving all three novel view synthesis metrics. These results support the effectiveness of updating proxies during refinement. With the additional computation from refinement, GLPA retains a $2.65\times$ speedup in attention computation per denoising step over dense attention on the FLUX backbone. On the SD backbone, GLPA without refinement improves reconstruction metrics but yields lower NVS quality than dense attention. The full model improves all five quality metrics over dense attention and both ablated variants, while retaining a $1.80\times$ speedup. These results support combining compact cross-view communication with recurrent updates for high-quality multi-view generation and reconstruction at lower computational cost.

\subsubsection{Pooling and Iteration Settings}
We further examine the selection of the pooling stride $s$ and the number of interaction rounds $R$. These two parameters control the spatial granularity of cross-view communication and the depth of iterative updates, respectively. Table~\ref{tab:settings_ablation} presents experimental results obtained by varying one parameter while keeping the other fixed.



\noindent\textbf{Pooling stride.}
A larger stride produces fewer proxy tokens, reducing the cost of
cross-view interaction but yielding a coarser representation.
With $R=2$ fixed, increasing $s$ from 2 to 4 or 8 reduces attention
runtime but degrades both generation and reconstruction quality.
These results suggest that retaining sufficient spatial detail
in the proxy representation is important for cross-view communication.

\noindent\textbf{Number of interaction rounds.}
With $s=2$ fixed, increasing $R$ from 1 to 2 improves all five
quality metrics, whereas a third round increases attention runtime
and degrades performance relative to two rounds.
We therefore adopt $s=2$ and $R=2$ as the default configuration,
which achieves the best generation and reconstruction quality
among the tested settings.
Qualitative comparisons of pooling strides and interaction rounds
are provided in the supplementary material.

\begin{table}[!t]
  \caption{Efficiency and Quality Comparison of Alternative Attention Designs on the FLUX Backbone.}
  \label{tab:flux_efficiency}
  \centering
  \footnotesize
  \renewcommand{\arraystretch}{1.2}
  \setlength{\tabcolsep}{2pt}
  \begin{tabular*}{\columnwidth}{@{\extracolsep{\fill}}l|ccccc@{}}
    \hline
    Methods &
    \makecell{Latency (ms)} &
    \makecell{Peak\\memory (G)} &
    LPIPS$\downarrow$ &
    SSIM$\uparrow$ &
    PSNR$\uparrow$ \\
    \hline
    Dense           & 26.61 & 0.828 & \textbf{0.101} & \textbf{0.918} & 25.84 \\
    PT-DiT\cite{wang2025pttiv}          & 35.41 & 1.913 & 0.145 & 0.901 & 24.39 \\
    Agent Attention\cite{han2024agentattention} & \textbf{21.74} & 1.986 & 0.137 & 0.909 & 24.84 \\
    GLPA            & 22.07 & \textbf{0.601} & 0.111 & 0.910 & \textbf{25.86} \\
    \hline
  \end{tabular*}
\end{table}

\begin{table}[!t]
  \caption{Efficiency Comparison of Attention Modules on the SD Backbone.}
  \label{tab:sd_efficiency}
  \centering
  \footnotesize
  \renewcommand{\arraystretch}{1.2}
  \setlength{\tabcolsep}{3pt}
  \begin{tabular*}{\columnwidth}{@{\extracolsep{\fill}}l|cccc@{}}
    \hline
    Methods & N-view & Resolution &
    \makecell{Latency (ms)} &
    \makecell{Peak memory (G)} \\
    \hline
    Dense & 8 & 512 & 14.92 & 0.244 \\
    Era3D\cite{li2024era3d} & 8 & 512 & \textbf{1.67} & 0.322 \\
    CTR3D\cite{luo2026ctr3d} & 8 & 512 & 9.20 & 0.325 \\
    GLPA & 8 & 512 & 5.02 & \textbf{0.156} \\
    \hline
  \end{tabular*}
\end{table}

\subsubsection{Comparison with Alternative Attention Designs}
We compare dense attention, PT-DiT, Agent Attention, and GLPA on the FLUX backbone under a unified training and evaluation protocol. All variants are trained for 2,000 optimization steps using the same 6,000 training samples and evaluated on the same 20 test samples, with eight views at a resolution of $512\times512$. We report the latency and peak allocated memory of a single attention-module forward pass, together with LPIPS, SSIM, and PSNR for novel view synthesis. As shown in Table~\ref{tab:flux_efficiency}, GLPA reduces attention latency and peak allocated memory compared with dense attention, achieving nearly identical PSNR but higher LPIPS and lower SSIM. Compared with PT-DiT and Agent Attention, GLPA achieves better performance across all three NVS metrics while requiring the least memory. Its latency is close to that of Agent Attention (22.07 vs. 21.74~ms), while its peak memory is substantially lower (0.601 vs. 1.986~G). Overall, these results demonstrate that GLPA achieves a favorable trade-off among attention latency, memory consumption, and NVS quality. On the SD backbone, Table~\ref{tab:sd_efficiency} presents an efficiency comparison. GLPA achieves a $2.97\times$ attention-module speedup over dense attention and reduces peak allocated memory from 0.244 to 0.156~G. While Era3D achieves the lowest latency, GLPA requires the least memory among the compared designs and is faster than CTR3D.


%% file: sec/5_conclusion.tex
\section{Conclusions}
In this paper, we have presented GLPA, an efficient attention mechanism for multi-view generation in single-image 3D reconstruction. GLPA preserves full-resolution within-view interactions while enabling efficient cross-view communication through compact global proxies. GLPA further incorporates iterative proxy refinement to improve cross-view consistency among generated views. This single-round global-local proxy interaction reduces attention runtime compared with dense attention. With iterative proxy refinement, the full GLPA configuration improves generation and reconstruction quality over the single-round variant. Extensive experiments on novel view synthesis and 3D reconstruction demonstrate the effectiveness of our method, producing consistent multi-view observations and complete 3D models.